\documentclass[11pt]{article}

\usepackage[preprint]{acl}

\usepackage{times}
\usepackage{latexsym}

\usepackage[T1]{fontenc}

\usepackage[utf8]{inputenc}

\usepackage{microtype}

\usepackage{inconsolata}

\usepackage{graphicx}

\usepackage{array}       
\usepackage{tabularx}    
\usepackage{booktabs}    
\usepackage{caption}     
\usepackage{amsmath} 
\usepackage{amsfonts}
\usepackage{subcaption}
\usepackage{algorithm}
\usepackage{algorithmic}
\usepackage{multirow}
\usepackage{makecell}
\usepackage{tcolorbox}
\usepackage{enumitem}
\usepackage{xcolor}     
\usepackage{tcolorbox}  

\title{Controllable Spoken Dialogue Generation: An LLM-Driven Grading System for K-12 Non-Native English Learners}
\author{
  \textbf{Haidong Yuan\textsuperscript{1}\thanks{Email: oseast@stu.pku.edu.cn}},
  \textbf{Haokun Zhao\textsuperscript{5}},
  \textbf{Wanshi Xu\textsuperscript{1}},
  \textbf{Songjun Cao\textsuperscript{2}},\\
  \textbf{Qingyu Zhou\textsuperscript{4}},
  \textbf{Long Ma\textsuperscript{2}},
  \textbf{Hongjie Fan\textsuperscript{3}}
  \\[4pt]
  \textsuperscript{1}Peking University,\quad
  \textsuperscript{2}Tencent,\\
  \textsuperscript{3}China University of Political Science and Law,\\
  \textsuperscript{4}Independent Researcher,\quad
  \textsuperscript{5}Fudan University
  \\[6pt]
  \small{\textbf{Correspondence to:} Hongjie Fan, Songjun Cao }
}

\begin{document}
\maketitle
\begin{abstract}
Large language models (LLMs) often fail to meet the pedagogical needs of K-12 English learners in non-native contexts due to a proficiency mismatch. To address this widespread challenge, we introduce a proficiency-aligned framework that adapts LLM outputs to learner abilities, using China's national curriculum (CSE) as a representative case. Our framework enables precise control over lexical complexity through a four-tier grading system, supported by a comprehensive suite of new resources: graded vocabulary lists and a multi-turn dialogue corpus.
Our core technical contribution is the \textbf{DDPO} algorithm,Diversity Driven Policy Optimization, a multi-turn GRPO-based approach designed to preserve dialogue diversity while holistically optimizing dialogue quality. This method significantly outperforms conventional approaches, achieving low out-of-vocabulary rates and high diversity while enhancing conversational naturalness and pedagogical value. While grounded in the CSE, our framework is designed for flexibility and can be readily adapted to other educational standards. Our models, data, and code will all be open-sourced, providing a scalable platform for personalized English speaking practice that effectively addresses the unique challenges faced by K-12 learners in non-immersive environments.
\end{abstract}
\begin{figure}[htb]
    \centering
    \includegraphics[width=1\linewidth]{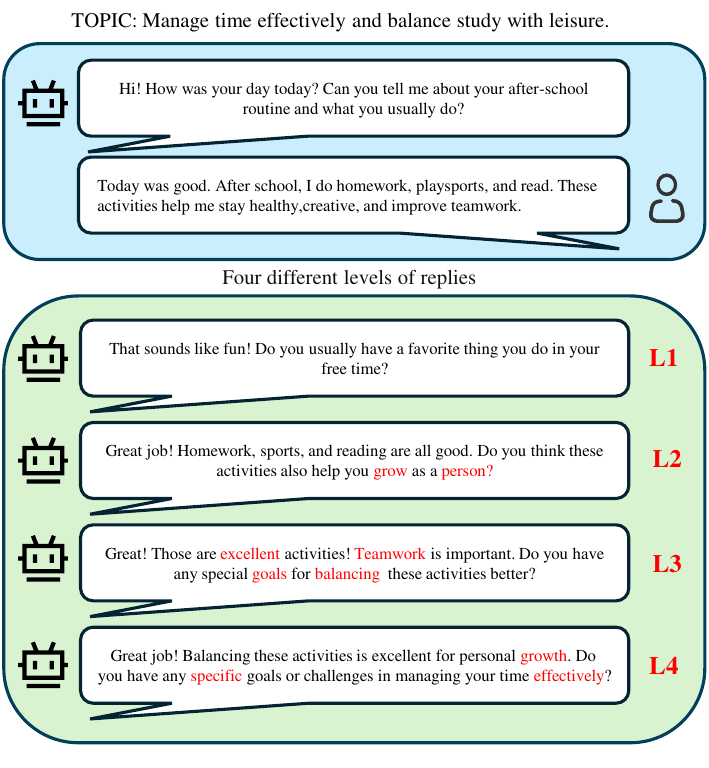}
    \caption{The figure shows examples of outputs at different proficiency levels for the same topic and input. Words in red indicate vocabulary specific to each level.}
    \label{fig: dialogue_example}
\end{figure}

\begin{figure*}[htbp]
    \centering
    \includegraphics[width=1\textwidth]{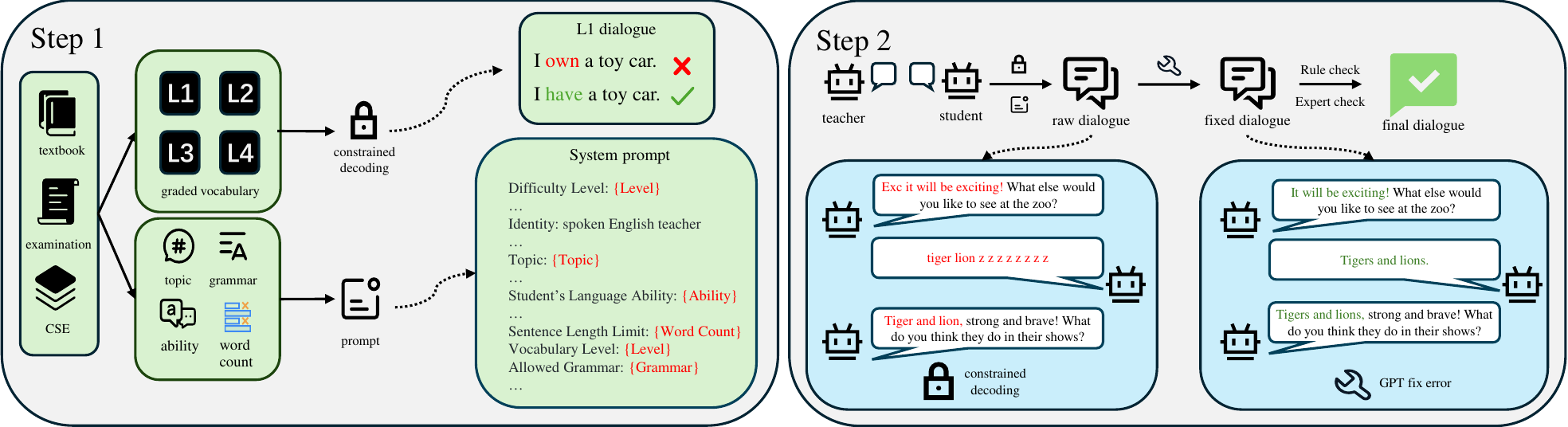}
    \caption{Data generation pipeline: Step 1, vocabulary, grammar, and proficiency information are extracted from textbooks and the CSE framework. The vocabulary list constrains decoding in Step 2, while other information is provided via prompts in this same step. Step 2, Two constrained-decoding language models then generate teacher-student dialogues using only the specified vocabulary. An unconstrained model corrects language errors, and finally, rule-based and human reviews select high-quality dialogues.}

    \label{fig:how to build data}
\end{figure*}
\section{Introduction}

For language learners, an ideal environment must provide ample opportunities for both input and output in the target language, as well as authentic interactive scenarios. However, primary and secondary school students in non-native English-speaking countries often lack authentic English communication environments, which severely limits their opportunities for oral practice and poses a significant barrier to improving their listening and speaking skills.

Recent advances in artificial intelligence have led to the widespread adoption of large language models (LLMs) in education, with particularly promising applications in language learning and oral practice~\cite{hagos2024recent, dong2024large, gao2023investigation}. Although numerous AI-powered English learning tools have emerged, most fail to fully address the specific needs of English education in these regions. Taking China as a representative case, these systems often lack fine-grained differentiation based on students’ cognitive development, vocabulary levels, and grammatical proficiency. This results in a mismatch between learning content and learners’ abilities, ultimately leading to suboptimal outcomes~\cite{li2024challenges}. According to the theory of the ``zone of proximal development"~\cite{danish2017observing}, the most effective learning occurs when students are exposed to language input that is just near their current level, which promotes both comprehension and progress. If the learning content is too complex, students may struggle with comprehension, undermining their motivation and learning efficacy; conversely, if the content is too simple, it lacks sufficient challenge and fails to stimulate further growth. So our system realizes dynamic adaptation of responses to students at different levels. For details, please refer to Figure~\ref{fig: dialogue_example}.

Furthermore, there are substantial differences in English language education frameworks across countries. For instance, while the Common European Framework of Reference for Languages (CEFR) is widely adopted and supported by extensive research and datasets~\cite{amiruddin2025empowering,Bacher_2024}, it differs considerably from the educational standards in China~\cite{peng2022aligning}, making it challenging to directly apply CEFR-based proficiency levels to local contexts. In contrast, China’s Standards of English Language Ability (CSE)~\cite{zhao2017calibrating, wu2021proficiency} are more closely aligned with the specific learning progression and educational context of Chinese students.

From a technical perspective, controlled text generation (CTG) methods—such as prompt engineering, constrained decoding, instruction tuning, and reinforcement learning (RL)~\cite{zhang2023survey}—offer effective techniques for guiding and constraining LLM outputs. Among these, reinforcement learning from human feedback (RLHF) has become the mainstream approach for aligning models with human preferences. For example, ChatGPT employs the Proximal Policy Optimization (PPO) algorithm~\cite{schulman2017proximal} to enhance dialogue consistency and alignment with human values. More recently, models such as DeepSeek-R1~\cite{guo2025deepseek} have utilized the PPO variant GRPO~\cite{shao2024deepseekmath} to optimize mathematical reasoning tasks, driving further advances in algorithms such as DAPO~\cite{yu2025dapo} and GRPO-LEAD~\cite{zhang2025grpo}. However, although reinforcement learning has been widely applied to enhance reasoning abilities, its application in aligning model outputs with language proficiency standards in educational contexts has received comparatively less attention.

\begin{figure*}[htbp]
    \centering
    \includegraphics[width=0.95\textwidth]{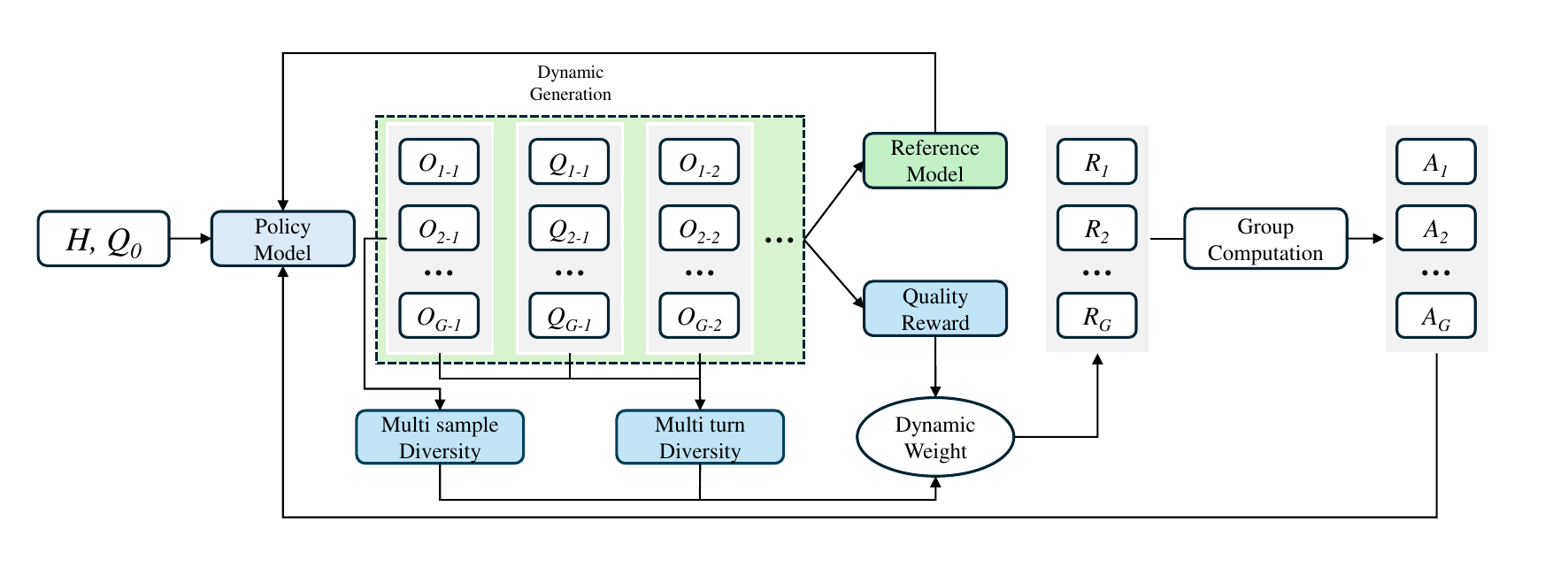}
    \caption{Overview of the Diversity Driven Policy Optimization (DDPO) framework. Given a fixed dialogue history $h$, we sample $G$ multi-turn trajectories. We utilize the $G$ first-turn responses to compute the \textit{Multi sample diversity}, and the sequential outputs within a single trajectory to compute the \textit{Multi-turn diversity}. Combined with a task-specific quality reward, these components are aggregated via dynamic weighting to determine the final reward, followed by advantage estimation and parameter updates.}
    \label{fig:DDPO}
\end{figure*}

To address these challenges, we propose an English proficiency grading and dialogue system tailored to the needs of K-12 learners in non-native contexts, using China's educational standards as a representative case study. Our main contributions are as follows:

\begin{itemize}
    \item \textbf{Diversity Driven Policy Optimization. (DDPO):} We introduce DDPO, a multi-turn reinforcement learning algorithm designed to mitigate the entropy collapse phenomenon often observed in GRPO during dialogue tasks. This approach effectively balances dialogue quality, response diversity, and reinforcement learning optimization objectives.
    
    \item \textbf{Comprehensive Graded Resources:} We construct a large-scale suite of pedagogical resources grounded in China's national curriculum and the China Standards of English (CSE). This includes hierarchically graded vocabulary lists, and a high-quality multi-turn dialogue corpus tailored for varying proficiency levels.

    \item \textbf{First CSE-Aligned Open-Source System:} We release the first open-source dialogue system fully aligned with the CSE framework. Extensive experiments demonstrate that our system significantly outperforms conventional prompting and supervised baselines, achieving superior proficiency control while maintaining high dialogue quality and instructional value.
\end{itemize}

\section{Related Work}

\subsection{Language Proficiency Frameworks}

The integration of language proficiency frameworks with LLMs has mainly focused on automated assessment and controlled generation—detecting proficiency levels in texts and guiding models to produce content matching specific abilities \cite{amiruddin2025empowering,lagutina2023text,redlich2024combining}. Internationally, the CEFR is the most widely used framework, classifying learners from A1 to C2. While research has explored the alignment between CEFR and  CSE \cite{coniam2022cefr,zhu2023effect}, LLM applications based on the CSE remain limited compared to those using CEFR.

Most English proficiency-annotated datasets, such as EFCAMDAT \cite{geertzen2013automatic}, CLC-FCE \cite{yannakoudakis2011new}, and CEFR-SP \cite{uchida-etal-2024-profiling}, are based on long texts and the CEFR framework. These datasets, typically consisting of multi-hundred-word passages, are suitable for overall difficulty assessment but not for training LLMs to generate short, natural conversational responses. TSCC \cite{caines-etal-2020-teacher} provides teacher-student dialogues with CEFR-level annotations, but its difficulty distribution is uneven and the content is more knowledge-based than conversational. DailyDialog \cite{li-etal-2017-dailydialog} offers natural and diverse conversations but lacks proficiency annotations, limiting its use for level-based modeling. Sentence-level datasets like CEFR-SP provide finer granularity but mainly contain isolated sentences, missing features of authentic dialogues such as phrases, single-word responses, and contextual references.

Therefore, there is a clear need for proficiency-graded dialogue corpora that span multiple difficulty levels and align with the CSE.

\subsection{Controlled Text Generation}

CTG methods, such as prompt engineering, constrained decoding, instruction tuning, and reinforcement learning, are widely used to regulate large language model outputs.

Prompt engineering constrains outputs via carefully designed prompts, and is often used for dialogue generation or task evaluation. For example, ~\citet{lin2024can} used GPT-4’s few-shot learning to generate fine-grained feedback on teacher errors, achieving accuracy surpassing human experts in some educational settings.

Constrained decoding restricts the generation process to meet specific requirements. For instance, \citet{tyen2022towards} applied CEFR-based vocabulary grading, \citet{qian2023user} increased the use of textbook vocabulary, and \citet{glandorf-etal-2025-grammar} encouraged target grammatical structures. However, such interventions may introduce errors or reduce output quality.

Instruction tuning encodes constraints as natural language instructions, as in the INSTRUCTCTG framework \cite{zhou2023controlled}, enabling models to learn constraints during training without modifying decoding. 

Reinforcement learning has also been applied for dialogue control. For example, \citet{du-etal-2024-rewarding} used stepwise rewards to improve dialogue completion, while \citet{cho2024deep} combined multiple reward functions to enhance response quality and controllability. These methods can model real-time learner feedback and dynamically optimize dialogue strategies.

\section{Dataset Construction}

To enable the model’s graded dialogue capabilities, a scientifically designed dataset classification is essential. Following Chinese English education standards, we divide students’ English proficiency into four levels (L1–L4), corresponding to Grades 1–4 in primary school, Grades 5–6 in primary school, junior high school, and senior high school, respectively. For each stage, we extract the relevant English proficiency requirements, focusing primarily on vocabulary. Accordingly, we construct a graded dialogue dataset based on level-specific vocabulary.
\begin{figure}[tb]
\centering
\begin{minipage}[t]{0.50\columnwidth}
\centering
\includegraphics[height=5cm, width=\linewidth, keepaspectratio]{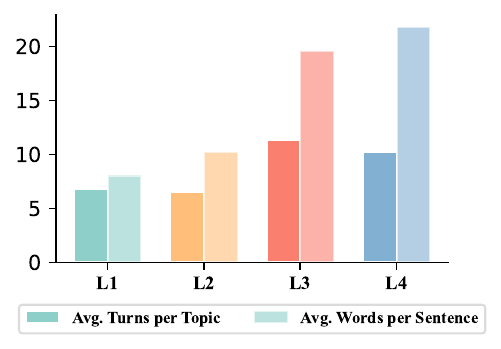}
\label{fig:bar_avg}
\end{minipage}
\hspace{0.11cm} 
\begin{minipage}[t]{0.46\columnwidth}
\centering
\includegraphics[height=4.5cm, width=\linewidth, keepaspectratio]{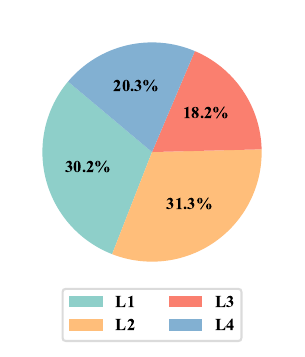}
\label{fig:pie_topic}
\end{minipage}
\caption{The train dataset covers dialogues from L1 to L4. (Left) Average turns per dialogue and words per turn for each grade; (Right) Topic distribution by grade.}
\label{fig:twopics}
\end{figure}

\subsection{Graded Dialogue Dataset}

According to the Chinese Ministry of Education, L1 students possess a vocabulary of only about 600 words, making it challenging to find suitable classroom dialogue texts at this level. To address this, our dialogue classification system prioritizes vocabulary difficulty. We first compile vocabulary lists from mainstream textbooks and categorize them into L1–L4, treating basic word forms (e.g., tense, plural) at the same level as their root forms.

We then extract dialogue topics and content from primary and secondary school exams and textbooks, ensuring that scenarios align with students’ communicative abilities at each level. Since most collected materials are not in a teacher-student dialogue format, we use constrained decoding with two Qwen2.5-32B to simulate such interactions, strictly limiting output to the allowed vocabulary. The resulting dialogues are in a question-and-answer format, with the teacher model guiding the student model based on the topic and proficiency level.

Although constrained decoding ensures that all vocabulary remains within the specified range, it may introduce grammatical errors, lexical mistakes, and meaningless repetitions. To address this, deepseek-v3 is employed to rewrite these dialogues, correcting errors without altering the content or vocabulary. The rewritten content is then subjected to another round of out-of-vocabulary (OOV) detection to ensure compliance, resulting in the final spoken dialogue corpus. The final output consists of a graded vocabulary list and a graded dialogue corpus. All proper nouns, such as personal and place names, are removed from the vocabulary list to prevent unreasonable OOV judgments during the final screening. The specific construction process is illustrated in Figure~\ref{fig:how to build data}.

\begin{table}[ht]
\centering
\footnotesize 
\setlength{\tabcolsep}{6pt} 
\begin{tabular}{lccc}
\toprule
& \textbf{Train} & \textbf{Validation} & \textbf{Test} \\
\midrule
Dialogue Turns & 6420 & 412 & 427 \\ 
Dialogue Topics & 786 & 51 & 41 \\
Words & 191894 & 12123 & 13356 \\ 
Avg. Turns per Topic & 8.17 & 8.08 & 10.41 \\
\bottomrule
\end{tabular}
\caption{Statistics of the dialogue datasets.}
\label{tab:dataset_stats}
\vspace{-2pt} 
\end{table}

Notably, dialogues at the junior and senior high school levels tend to be longer, resulting in a relatively smaller number of topics compared to the primary school level. Detailed distributions of topics and turns are illustrated in Figure~\ref{fig:twopics} and Table~\ref{tab:dataset_stats}.

\section{Methodology}

\begin{table}[t]
    \centering
    \resizebox{\linewidth}{!}{
        \begin{tabular}{@{}l c c c c c@{}} 
        \toprule
        \multirow{2}{*}{\textbf{Model}} & \textbf{Human} & \multicolumn{4}{c}{\textbf{Judge Models}} \\ 
        \cmidrule(l){3-6} 
        & \textbf{Score} & \textbf{GPT5.1} & \textbf{Gemini} & \textbf{DouBao} & \textbf{Hunyuan-T1} \\
        \midrule
        GPT-5.1     & 4.47$^{(4)}$ & 4.07$^{(4)}$ & 3.93$^{(4)}$ & 3.99$^{(4)}$ & 4.24$^{(4)}$ \\
        Gemini-3.0  & 4.63$^{(1)}$ & 4.37$^{(1)}$ & 4.37$^{(1)}$ & 4.21$^{(2)}$ & 4.37$^{(2)}$ \\
        Doubao-1.6      & 4.48$^{(3)}$ & 4.16$^{(3)}$ & 4.02$^{(3)}$ & 4.15$^{(3)}$ & 4.26$^{(3)}$ \\
        HunYuan-turbos     & 4.61$^{(2)}$ & 4.35$^{(2)}$ & 4.26$^{(2)}$ & 4.36$^{(1)}$ & 4.40$^{(1)}$ \\
        \bottomrule
        \end{tabular}
    }
    \caption{Comparison of average scores and rankings (in parentheses) between human evaluation and various LLM-based judges. The rankings from judge models align closely with human evaluation.}
    \label{tab:model_vs_human}
\end{table}
\subsection{Entropy Collapse}
In the iterative process of Group Relative Policy Optimization (GRPO), the policy model tends to disproportionately reinforce specific token paths that exhibit marginal advantages during intra-group sampling. This causes the probability mass to rapidly concentrate on local optima, leading to a sharp decline in generation diversity---a phenomenon we term \textit{Entropy Collapse}. In dialogue tasks, this results in severe homogenization (both semantic and syntactic) across multiple samples and multi-turn interactions, significantly degrading user experience. Examples of Entropy Collapse are presented in Appendix G.

\subsection{Diversity Driven Policy Optimization}
To address the entropy collapse, we propose \textbf{Diversity Driven Policy Optimization (DDPO)}. DDPO integrates a user simulator for multi-turn trajectory generation and introduces a dynamic diversity mechanism to balance exploration and exploitation, show in Figure~\ref{fig:DDPO}.

\paragraph{Multi-turn Sampling}
Given an initial prompt $x$, we generate $G$ independent dialogue sessions by interacting with a user simulator $\mathcal{U}$ using the policy $\pi_\theta$. The trajectory for the $i$-th group, denoted as $\tau_i$, consists of a sequence of $K$ turns:
\begin{equation*}
\begin{aligned}
\tau_i &= \{(u_{i,1}, a_{i,1}), \dots, (u_{i,K}, a_{i,K})\} \\
\text{s.t.} \quad a_{i,k} &\sim \pi_\theta(\cdot|h_{i,k}), \quad u_{i,k+1} \sim \mathcal{U}(\cdot|h_{i,k}, a_{i,k})
\end{aligned}
\end{equation*}
where $h_{i,k}$ represents the dialogue history for group $i$ at turn $k$. The terms $u_{i,k}$ and $a_{i,k}$ denote the user utterance and model response at turn $k$, respectively. Note that all groups are initialized with the same prompt, i.e., $u_{i,1} = x$.

\paragraph{Single-turn Diversity (Cross-Sample)}
The first turn ($k=1$) is prone to mode collapse. We calculate the negative average Rouge-L F1 score of response $a_{i,1}$ against other samples in the same group. To prevent meaningless gibberish caused by excessive diversity maximization, we apply a clipping threshold $\gamma$:
\begin{equation*}
\resizebox{\linewidth}{!}{$
R_{\text{sgl}}(a_{i,1}) = - \max \left( \frac{1}{G-1} \sum_{j \neq i} \text{RougeL}(a_{i,1}, a_{j,1}), \ \gamma \right)
$}
\end{equation*}

\paragraph{Multi-turn Diversity (Cross-Time)}
For subsequent turns ($k>1$), we address contextual repetition. We penalize the token overlap ratio of the current response $a_{i,k}$ with the current user input $u_{i,k}$ and the previous assistant output $a_{i,k-1}$:
\begin{equation*}
R_{\text{mul}}(a_{i,k}) = - \left( \frac{|a_{i,k} \cap u_{i,k}|}{|a_{i,k}|} + \frac{|a_{i,k} \cap a_{i,k-1}|}{|a_{i,k}|} \right)
\end{equation*}
where $|S|$ denotes the unique token count and $\cap$ denotes token set intersection.

\paragraph{Dynamic Reward Composition}
We incorporate task-specific quality rewards (e.g., correctness or reward model scores), denoted as $R_{\text{qual}}$. To effectively balance exploration and exploitation, we employ a dynamic weighting schedule based on the global training step $S$. The composite reward $R_{i,k}$ for the $k$-th turn of the $i$-th sample is:
\begin{equation*}
\begin{aligned}
R_{i,k} =\; & \lambda_{\text{qual}} \cdot R_{\text{qual}}(a_{i,k}) \\
& + \lambda_{\text{sgl}} \cdot R_{\text{sgl}}(a_{i,1}) \\
& + \lambda_{\text{mul}} \cdot R_{\text{mul}}(a_{i,k})
\end{aligned}
\end{equation*}
The coefficients $\lambda$ are independent weights for each reward, intended to adjust the influence of each component during different training phases.

\paragraph{Optimization Objective}
We adopt token-level averaging to eliminate length bias. The DDPO objective function is defined as:
\begin{equation*}
\resizebox{0.9\linewidth}{!}{$
\begin{aligned}
    \mathcal{J}_{\mathrm{DDPO}}(\theta) = \mathbb{E}_{\substack{q \sim P(Q) \\ \{\tau_i\}_{i=1}^G \sim \pi_{\theta_{\mathrm{old}}}}} \Bigg[ 
     \frac{1}{Z} \sum_{i=1}^G \sum_{k=1}^{K} \sum_{t=1}^{|a_{i,k}|} 
    \min \left( \mathcal{L}^{\text{clip}}_{i,k,t} \right) \Bigg]
\end{aligned}
$}
\end{equation*}
where $Z = \sum_{i=1}^G \sum_{k=1}^K |a_{i,k}|$ is the total token count in the batch. The clipped surrogate loss is:
\begin{equation*}
\begin{aligned}
    \mathcal{L}^{\text{clip}}_{i,k,t} &= \left( \rho_{i,k,t} \hat{A}_{i,k}, \ \text{clip}(\rho_{i,k,t}, 1-\epsilon, 1+\epsilon) \hat{A}_{i,k} \right) \\
    \rho_{i,k,t} &= \frac{\pi_\theta(a_{i,k,t} | h_{i,k}, a_{i,k,<t})}{\pi_{\theta_{\mathrm{old}}}(a_{i,k,t} | h_{i,k}, a_{i,k,<t})}
\end{aligned}
\end{equation*}
Here, $\hat{A}_{i,k}$ is the standardized advantage for the $k$-th turn, computed relative to the group statistics of that specific turn:
\begin{equation*}
\hat{A}_{i,k} = \frac{R_{i,k} - \mu_k}{\sigma_k + \delta}
\end{equation*}
where $\mu_k = \frac{1}{G}\sum_{j=1}^G R_{j,k}$ and $\sigma_k = \text{std}(\{R_{j,k}\}_{j=1}^G)$ represent the mean and standard deviation of rewards for turn $k$ across the group, and $\delta$ is a small constant for numerical stability.

\section{Evaluation Metrics}

A spoken language practice system must satisfy three core objectives: (1) \textbf{Comprehensibility}, ensuring vocabulary remains within the learner's proficiency level; (2) \textbf{Diversity}, avoiding monotonous or repetitive outputs; and (3) \textbf{Educational Quality}, ensuring the dialogue is error-free, informative, and capable of guiding the conversation effectively. Accordingly, we establish a three-dimensional evaluation framework covering vocabulary constraints, diversity, and dialogue quality.

\subsection{Vocabulary}

Vocabulary evaluation relies on pre-constructed graded vocabulary lists. During the dialogue, the system instructs the model to adopt a specific persona and difficulty level based on the scenario, restricting usage to the assigned vocabulary tier. A \textit{vocabulary violation} is recorded if the model generates tokens exceeding the learner's proficiency level (e.g., using L4 words in an L3 session). However, proper nouns (e.g., names, locations) and out-of-vocabulary (OOV) terms previously introduced in the dialogue history are exempt. This exception ensures naturalness, as proper nouns may not be in the predefined lists, and reusing established OOV terms is essential for coherent interaction.

\subsection{Diversity Evaluation}
Diversity is assessed through both cross-sample and cross-turn dimensions. Specifically, for a fixed dialogue history, we generate eight independent multi-turn trajectories. We compute the average Rouge-L score across the first turns of these samples (inter-sample diversity) and the Rouge-L score across the generated multi-turn teacher-student interactions (intra-session diversity). The final diversity metric is calculated as the 1:1 weighted sum of these two scores.

\subsection{Dialogue Quality Evaluation}
Unlike general-purpose dialogue systems, evaluation in spoken practice must align with specific educational objectives~\cite{mendonca-etal-2024-soda}. Beyond contextual relevance, the model must actively encourage learner participation. We instruct the model to adopt a teacher persona employing a ``Response + Expansion + Question'' strategy: acknowledging the student's input, expanding on the idea, and posing a follow-up question based on the current topic~\cite{bowden-etal-2024-active}. This ensures the dialogue remains pedagogically valuable, engaging, and logically coherent.

We employ five criteria to evaluate dialogue quality:
\begin{itemize}
    \item \textbf{Topic Relevance:} The response must be closely aligned with the designated dialogue topic.
    \item \textbf{Task Completion:} The model must fulfill specific constraints in the prompt, such as length, theme, or response format.
    \item \textbf{Semantic Richness:} The response should be informative and varied, avoiding generic or hollow content.
    \item \textbf{Topic Guidance:} The output should effectively guide the student to think critically and express themselves.
\end{itemize}

\begin{table}[t]
\centering
\resizebox{\linewidth}{!}{
    \begin{tabular}{@{} l c c c c c c c @{}} 
    \toprule
    \multirow{2}{*}{\textbf{Model}} & 
    \multicolumn{2}{c}{\textbf{Objective}} & 
    \multicolumn{5}{c}{\textbf{Quality (1-5)}} \\
    \cmidrule(lr){2-3} \cmidrule(l){4-8}
     & \textbf{Voc.} $\downarrow$ & \textbf{Div.} $\uparrow$ & \textbf{Rel.} $\uparrow$ & \textbf{Task} $\uparrow$ & \textbf{Rich.} $\uparrow$ & \textbf{Gui.} $\uparrow$ & \textbf{Avg.} $\uparrow$ \\
    \midrule
    
    \multicolumn{8}{l}{\textit{Qwen2.5-7B Series}} \\
    \quad Base & 41.69 & 0.66 & 4.97 & 3.21 & 3.58 & 4.50 & 4.06 \\
    \quad + Prompt & 45.43 & 0.64 & 4.97 & 3.41 & \textbf{3.70} & 4.54 & 4.15 \\
    \quad + Decoding & \textbf{9.36} & 0.65 & 4.95 & 3.10 & 3.47 & 4.49 & 4.00 \\
    \quad + SFT & 23.18 & \textbf{0.76} & \textbf{4.98} & 3.33 & 3.51 & 4.47 & 4.07 \\
    \quad + \textbf{DDPO} & \textbf{9.36} & 0.72 & 4.96 & \textbf{3.63} & 3.64 & \textbf{4.87} & \textbf{4.30} \\
    
    \midrule
    
    \multicolumn{8}{l}{\textit{Llama3.1-8B Series}} \\
    \quad Base & 55.26 & 0.72 & 4.98 & 3.58 & \textbf{3.90} & 4.48 & 4.23 \\
    \quad + SFT & 19.90 & \textbf{0.74} & \textbf{4.99} & 3.60 & 3.48 & 4.51 & 4.14 \\
    \quad + \textbf{DDPO} & \textbf{6.79} & 0.72 & 4.94 & \textbf{3.80} & 3.79 & \textbf{4.84} & \textbf{4.34} \\
    
    \bottomrule
    \end{tabular}
}
\caption{Main results on the spoken practice evaluation set. \textbf{Voc.}: Vocabulary Violation Rate; \textbf{Div.}: Diversity Score; \textbf{Rel.}: Topic Relevance; \textbf{Rich.}: Semantic Richness; \textbf{Gui.}: Topic Guidance. The best results in each group are highlighted in \textbf{bold}.}
\label{tab:result}
\end{table}

\paragraph{LLM-based Evaluation}
Human evaluation is susceptible to subjective bias and inter-rater variability, making standardization difficult. Furthermore, large-scale manual annotation is prohibitively expensive. However, recent work by \citet{long2024evaluating} demonstrates that LLMs possess strong capabilities in natural language understanding and pattern recognition suitable for educational data assessment. 

To validate the feasibility of LLM-based evaluation, we conducted a preliminary study using a mainstream closed-source LLM. We compiled a test set of 100 samples and collected ratings from four annotators with graduate-level English proficiency. We then prompted the LLM to perform the same assessment. As shown in Table~\ref{tab:model_vs_human}, the LLM's ranking alignment with human annotators is substantial, particularly in identifying lower-quality models where the rankings were identical. Based on these results, we selected \textbf{GPT-5.1} as the judge model for our experiments.

\section{Experiments}

We investigate three distinct approaches for training spoken dialogue agents, ranging from inference-time constraints to parameter-efficient tuning and reinforcement learning.

\subsection{Non-fine-tuning Methods}
We first explore methods that do not require parameter updates. 
The first approach is \textbf{Prompting}, where the complete list of allowed vocabulary is explicitly included in the system prompt, relying on the model's in-context learning capability to select appropriate words. 
The second approach is \textbf{Constrained Decoding}, which enforces strict output control. Following ~\citet{tyen2022towards}, we implement a dynamic constraint mechanism during inference. At each decoding step, the system maintains a set of valid next-tokens based on the current prefix and the allowed vocabulary, explicitly masking out OOV terms, special symbols, and tokens from other languages.

\subsection{Supervised Fine-Tuning (SFT)}

Supervised Fine-Tuning serves as a robust baseline for task-specific adaptation. We employ \textbf{Qwen2.5-7B} and \textbf{Llama3.1-8B} as backbone models to evaluate the generalizability of our approach across different architectures. To ensure sufficient adaptation, we construct a training set comprising 1,000 dialogue turns for each proficiency level. This process aims to align the models with the specific interactive requirements of spoken language practice.

\subsection{Reinforcement Learning (RL)}

In this task, the quality reward ($R_{\text{qual}}$) is primarily designed as a rule-based vocabulary grading mechanism (detailed in Appendix A. Employing this reward in isolation constitutes our baseline method, referred to as \textbf{GRPO}. Additionally, we investigate the impact of incorporating a model-based quality scorer, which is trained on 1,700 pairs of human-annotated preference data. Regarding our proposed \textbf{DDPO}, we conduct ablation studies to isolate the specific contributions of its two core diversity components: the multi-turn ($R_{\text{mul}}$) and single-turn ($R_{\text{sgl}}$) diversity rewards. The results of these ablation studies are presented in Table \ref{tab:ablation_results}.

\begin{table}[t]
\centering
\resizebox{\linewidth}{!}{
    \begin{tabular}{@{} l c c c c c c c @{}} 
    \toprule
    \multirow{2}{*}{\textbf{Method}} & 
    \multicolumn{2}{c}{\textbf{Objective}} & 
    \multicolumn{5}{c}{\textbf{Quality (1-5)}} \\
    \cmidrule(lr){2-3} \cmidrule(l){4-8}
     & \textbf{Voc.} $\downarrow$ & \textbf{Div.} $\uparrow$ & \textbf{Rel.} $\uparrow$ & \textbf{Task} $\uparrow$ & \textbf{Rich.} $\uparrow$ & \textbf{Gui.} $\uparrow$ & \textbf{Avg.} $\uparrow$ \\
    \midrule
    
    Llama-3.1-8B & 55.26 & 0.72 & 4.98 & 3.58 & \textbf{3.90} & 4.48 & 4.23 \\
    
    GRPO & \textbf{2.57} & 0.39 & 4.90 & 3.40 & 3.28 & 3.81 & 3.84 \\

    \quad + $R_{\text{model}}$ & 13.81 & 0.53 & \textbf{4.99} & \textbf{3.89} & 3.75 & 4.71 & 4.33 \\

    \quad + $R_{\text{mul}}$ & 6.56 & 0.61 & 4.94 & 3.55 & 3.35 & \textbf{4.95} & 4.20 \\
    
    \quad + $R_{\text{sgl}}$ & 3.27 & 0.64 & 4.48 & 3.07 & 3.27 & 4.14 & 3.74 \\
    
    \quad +$R_{\text{model}}$+ $R_{\text{mul}}$ & 8.90 & 0.61 & 4.97 & 3.61 & 3.58 & 4.73 & 4.22 \\

    \midrule
    
    \textbf{DDPO (Ours)} & 6.79 & \textbf{0.72} & 4.94 & 3.80 & 3.79 & 4.84 &\textbf{4.34} \\
    
    \bottomrule
    \end{tabular}
}
\caption{Ablation study results. DDPO strikes a favorable trade-off, ensuring both vocabulary adherence and diversity while delivering high-quality dialogue.}
\label{tab:ablation_results}
\end{table}

\section{Analysis}

\subsection{Main Results Analysis}
Table~\ref{tab:result} presents the performance of different approaches across Qwen2.5-7B and Llama3.1-8B architectures.

\paragraph{Limitations of Non-Fine-Tuning Methods}
Standard LLMs struggle to adhere to strict vocabulary constraints through prompting alone. As shown in the Qwen series, the \textbf{Prompting} method fails to effectively reduce the Vocabulary Violation rate (45.43\%), performing similarly to the Base Model (41.69\%). While \textbf{Constrained Decoding} successfully minimizes vocabulary violations (9.36\%), it severely hampers generation quality. The rigid constraints can disrupt the model's language modeling probability, leading to grammatical mistakes and unforeseen vocabulary errors, which accounts for the residual 9.36\% violation rate. The average quality score drops to 4.00. indicating that rigid decoding constraints force the model into unnatural or error phrasing.

\paragraph{Effectiveness of SFT}
Supervised Fine-Tuning (\textbf{SFT}) significantly improves upon the base model, achieving the highest Diversity scores (0.76 for Qwen, 0.74 for Llama). However, SFT alone is insufficient for strict pedagogical constraints. For instance, the Llama3.1-SFT model still retains a vocabulary violation rate of 19.90\%, which is suboptimal for learners requiring precise difficulty matching.

\paragraph{Superiority of DDPO}
Our proposed \textbf{DDPO} method achieves the best trade-off between constraints and quality. It matches the strict vocabulary control of Constrained Decoding (e.g., 9.36\% on Qwen, 6.79\% on Llama) while surpassing SFT in overall dialogue quality (Average 4.30 and 4.34, respectively). Meanwhile, DDPO maintains relatively good diversity.

\begin{figure}[t]
    \centering
    \begin{subfigure}{0.9\linewidth}
        \centering
        \includegraphics[width=\linewidth]{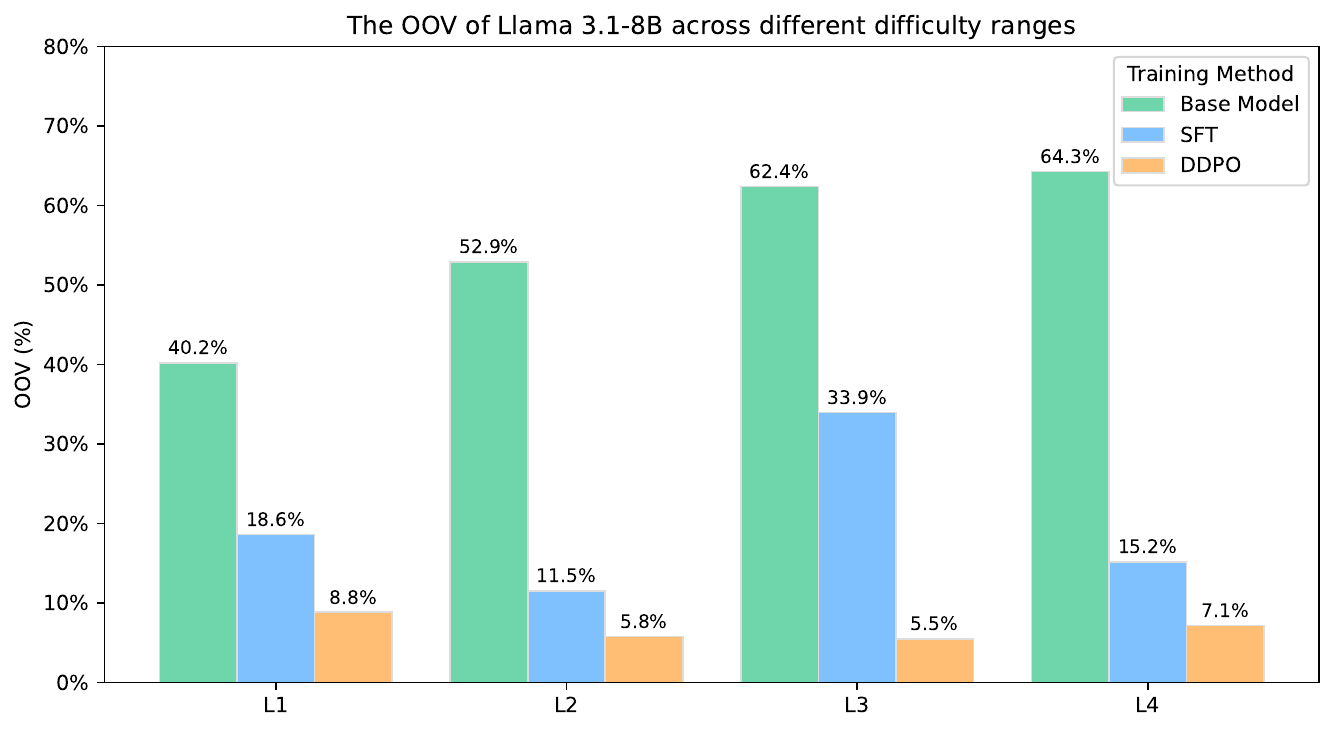}
    \end{subfigure}
    \caption{Proportion of dialogues with OOV. Compared with the base model and SFT method, the DDPO method significantly reduces the OOV at each level.}
    \label{fig:out-of-level}
    
\end{figure}

\paragraph{Comparison of Different Training Methods}
As shown in Figure~\ref{fig:out-of-level}, the OOV ratio of Llama-3.1-8B is extremely high at every level. Moreover, as the difficulty level increases, the proportion of OOV gradually rises, with the highest reaching 64.3\% at L4. After SFT  training, the OOV ratios of all levels—except Level 3 (L3)—drop sharply, yet they still remain above 10\%. Following DDPO training, the OOV ratios of all levels are reduced to below 10\%, and the optimization effect on L3 is particularly significant.

\subsection{Ablation Study}
We conducted an ablation study on the Llama3.1-8B model to investigate the specific contribution of each reward component within the DDPO framework. The detailed results are presented in Table~\ref{tab:ablation_results}.

\paragraph{Impact of Hard Constraints (GRPO Baseline)}
The baseline GRPO model, optimized solely with vocabulary constraints, achieves the lowest vocabulary violation rate (2.57\%). However, this comes at a significant cost to diversity (0.39) and semantic richness (3.28), resulting in the lowest average quality score (3.95). This confirms that optimizing for constraints alone leads to repetitive and safe but unengaging responses; specifically, the GRPO algorithm, when relying solely on rule-based quality rewards, exhibits a distinct \textbf{entropy collapse} phenomenon. Incorporating a model-based reward scorer effectively alleviates quality issues; however, diversity remains suboptimal (0.53), and we observe a moderate increase in OOV violations.

\paragraph{Trade-offs in Diversity and Quality Rewards}
Adding the multi-turn diversity reward ($+R_{\text{mul}}$) significantly recovers diversity (0.61), proving essential for preventing repetitive loops. Interestingly, while the single-turn diversity reward ($+R_{\text{sgl}}$) improves diversity (0.64), it negatively impacts dialogue quality, leading to a lower average score (3.74). We hypothesize that in its pursuit of diversity given a fixed dialogue history, the model may resort to incoherent word usage or random stacking, thereby degrading semantic quality. When both the model-based and multi-turn rewards are integrated, diversity does not increase further, but dialogue quality is notably restored.

\paragraph{The Balance of DDPO}
The full DDPO configuration effectively integrates these conflicting objectives. By combining vocabulary constraints with both diversity rewards and the model reward, DDPO maintains a low violation rate (6.79\%) and high diversity (0.72) while achieving the \textbf{best} average quality score (4.34). This demonstrates that the DDPO method effectively optimizes target rewards while successfully preventing entropy collapse.

\section{Conclusion}

In this work, we address the critical need for proficiency-aligned oral English practice for K-12 students in non-native contexts. Using China’s educational standards as a representative framework, we constructed comprehensive graded vocabulary, and dialogue datasets to facilitate this task.

Our primary contribution is the proposal of \textbf{DDPO (Diversity Driven Policy Optimization)}, a reinforcement learning framework designed to balance strict pedagogical constraints with dialogue diversity. Our experiments demonstrate that while standard methods like Constrained Decoding or SFT struggle to simultaneously satisfy vocabulary limitations and generation diversity, DDPO effectively aligns Large Language Models with these conflicting objectives. It achieves precise proficiency control comparable to hard constraints while maintaining the naturalness and semantic richness of open-ended generation.

This approach proves robust across different model architectures (Qwen and Llama), highlighting its potential as a generalizable solution for educational dialogue systems. In future work, we plan to extend our framework to other languages and curricula, and explore the integration of multi-modal capabilities, such as speech recognition and prosody analysis, to provide a more immersive language learning experience.

\section*{Limitations}

Despite the promising results of DDPO in the field of spoken language education, we acknowledge several limitations in our current work:

\begin{itemize}
    \item \textbf{Reliance on User Simulators:} Our training and evaluation primarily rely on a user simulator. While designed to mimic learner behavior, it cannot fully capture the unpredictability, emotional nuances, and diverse error patterns of real human students. Future research should incorporate methods based on user modeling to validate effectiveness in real-world scenarios.

    \item \textbf{Computational Overhead:} The DDPO framework requires sampling multiple trajectories ($G$ groups) and calculating diversity rewards across them. Compared to SFT or basic GRPO implementations, this incurs a higher computational cost during the training phase.

    \item \textbf{Model Scale and Generalization:} Our experiments were conducted on the Llama-3.1-8B model. While DDPO has demonstrated effectiveness at this scale, we have not yet explored its performance on larger-scale models (e.g., 70B parameters) or different model architectures.

    \item \textbf{Static Vocabulary Constraints:} The graded vocabulary lists used in the current educational setting are static. Consequently, they may not fully accommodate compound words or morphological variations (e.g., prefixes and suffixes), potentially treating valid derivations as unknown words. Future work needs to investigate the impact of dynamic graded vocabularies.
\end{itemize}

\bibliography{latex/custom}

\appendix

\section{Vocabulary Reward}
\label{sec:appendix}
\begin{algorithm}[tb]
\small
\caption{Vocabulary-Constraint Quality Reward Function}
\label{alg:accuracy-reward}
\begin{algorithmic}[1]
\REQUIRE $\mathcal{R}$: list of responses; $\mathcal{L}$: target levels (L1-L4); $\mathcal{V}$: graded vocabulary dictionary
\ENSURE $S$: list of reward scores

\FOR{each $(r, l)$ in $(\mathcal{R}, \mathcal{L})$}
    \STATE \textbf{Initialization:} $N_{words} \leftarrow 0$, $N_{target} \leftarrow 0$, $Flag_{violation} \leftarrow \text{False}$
    \STATE \textbf{Preprocessing:} Tokenize $r$, remove punctuation.
    
    \FOR{each word $w$ in $r$}
        \IF{$w$ is entity (NER), number, or spoken filler}
            \STATE \textbf{continue}
        \ENDIF
        \STATE $w_{lemma} \leftarrow \text{Lemmatize}(w)$
        \STATE $N_{words} \leftarrow N_{words} + 1$
        
        \IF{$w_{lemma} \notin \mathcal{V}$ or $\text{Level}(w_{lemma}) > l$}
            \STATE $Flag_{violation} \leftarrow \text{True}$
        \ENDIF
        \IF{$\text{Level}(w_{lemma}) == l$}
            \STATE $N_{target} \leftarrow N_{target} + 1$
        \ENDIF
    \ENDFOR

    \STATE \textbf{Hard Constraint Check:}
    \IF{sentences count $\le 1$ \OR '?' $\notin r$ \OR count('?') $\ge 2$ \OR contains non-English}
        \STATE $score \leftarrow 0.0$
    \ELSE
        \STATE Define length range $[min, max]$ based on $l$:
        \STATE \quad L1: $[10, 15]$, L2: $[10, 20]$, L3/L4: $[20, 30]$
        
        \IF{$N_{words} \in [min, max]$ \AND \textbf{not} $Flag_{violation}$}
            \IF{$l == \text{L1}$}
                \STATE $score \leftarrow 0.8$
            \ELSE
                \STATE $score \leftarrow 0.5 + \min(N_{target} \times 0.15, 2.0)$
            \ENDIF
        \ELSE
            \STATE $score \leftarrow 0.2$ \COMMENT{Soft penalty for length/vocab violation}
        \ENDIF
    \ENDIF
    \STATE Append $score$ to $S$
\ENDFOR
\RETURN $S$
\end{algorithmic}
\end{algorithm}

\section{Main Training and Testing Parameters}

\subsection*{Training Settings}

\textbf{Qwen}

\begin{itemize}
    \item \textbf{SFT:} Full-parameter fine-tuning, learning rate of $2 \times 10^{-5}$, warm-up ratio of 0.05, trained for 2 epochs.
    \item \textbf{DDPO:} Learning rate of $3.0 \times 10^{-6}$, warm-up ratio of 0.1, trained for 1 epochs.
\end{itemize}

\textbf{LLaMA}

\begin{itemize}
    \item \textbf{SFT:} Full-parameter fine-tuning, learning rate of $2 \times 10^{-5}$, warm-up ratio of 0.05, trained for 2 epochs.
    \item \textbf{DDPO:} Learning rate of $3.0 \times 10^{-6}$, warm-up ratio of 0.1, trained for 1 epochs.
\end{itemize}

\subsection*{Testing Settings}

For both the generation model and the scoring model, the temperature is set to 0.7 and the beam size is set to 1. 
For each input, eight samples are collected to be used in diversity calculation.

\section{Proficiency Levels}
\label{app:proficiency}

Table \ref{tab:proficiency_levels} details the four proficiency levels (L1--L4) defined in our dataset, ranging from basic recognition to complex logical expression.

\begin{table}[h!]
\centering
\small
\renewcommand{\arraystretch}{1.4}
\begin{tabularx}{\linewidth}{@{}l X@{}}
\toprule
\textbf{Level} & \textbf{Description} \\
\midrule
\textbf{L1} & Can understand and use simple words and phrases in daily life. Recognizes basic information (colors, numbers, greetings), reads short texts with pictures, introduces oneself, and writes basic words. Uses simple grammar for basic communication. \\
\textbf{L2} & Can understand slow, simple conversations and stories with support. Reads short texts to extract main ideas. Describes familiar people and events; expresses feelings; uses basic tenses and polite expressions. Pronunciation is mostly correct. \\
\textbf{L3} & Can understand complex daily conversations and broadcasts. Reads and analyzes short articles to infer logic. Communicates coherently on familiar topics, expresses opinions, and participates in discussions using a range of tenses and structures. \\
\textbf{L4} & Can understand multi-person conversations, news, and lectures. Reads various text types to distinguish facts from opinions. Expresses ideas fluently and logically, adapts to different situations, and uses complex grammar and vocabulary accurately. \\
\bottomrule
\end{tabularx}
\caption{Detailed descriptions of the four English proficiency levels used in the task.}
\label{tab:proficiency_levels}
\end{table}

\section{Dialogue Model System Prompt}
\label{app:model_prompt}

The system prompt used for the dialogue agent (Anna) is presented below. Variables enclosed in curly braces (e.g., \texttt{\{level\}}) are dynamically filled based on the specific user profile and task configuration.

\begin{tcolorbox}[colback=gray!5!white, colframe=gray!75!black, title=\textbf{System Prompt for Dialogue Agent}]
\textbf{Configuration}
\begin{itemize}[nosep]
    \item \textbf{Difficulty Level:} \texttt{\{level\}}
    \item \textbf{Role:} Anna (24-year-old spoken English teacher)
    \item \textbf{Student Ability Reference:} \texttt{\{Ability\}}
\end{itemize}

\textbf{Role Goals}
\begin{enumerate}[leftmargin=*, nosep]
    \item Engage in an English conversation with a student at \texttt{\{grade\_level\}} on the topic: \texttt{\{topic\}}.
    \item In each turn, summarize/expand on the previous input and ask a follow-up question.
\end{enumerate}

\textbf{Task Constraints}
\begin{enumerate}[leftmargin=*, nosep]
    \item \textbf{Length \& Vocab:} Keep answers within \texttt{\{word\_count\}} words. Use vocabulary suitable for \texttt{\{level\}}.
    \item \textbf{Grammar Constraint:} You may \textit{only} use the following grammar: \texttt{\{grammar\}}. Do not use unlisted grammar.
    \item \textbf{Adaptability:} Match the student's ability level---neither too difficult nor too simple.
\end{enumerate}

\textbf{Output Requirements}
\begin{enumerate}[leftmargin=*, nosep]
    \item \textbf{Structure:} Response must include a \textit{Reaction} (respond/expand) and a \textit{Question} (open-ended follow-up).
    \item \textbf{Guidance:} If the conversation strays, guide the student back to the topic.
\end{enumerate}
\end{tcolorbox}

\section{Dialogue Quality Evaluation Prompt}

\label{app:eval_prompt}

Table \ref{tab:eval_prompt} presents the detailed system prompt used for the model-based subjective evaluation. The evaluator (e.g., GPT-4) is instructed to act as a rigorous dialogue quality assessor using a 5-point Likert scale across eight specific dimensions. \textbf{Crucially, these evaluation criteria are identical to the guidelines provided to our human annotators to ensure consistency.}

\begin{table*}[h!]
\centering
\small
\begin{tcolorbox}[colback=gray!5!white, colframe=gray!75!black, title=\textbf{System Prompt for Subjective Evaluation}]
You are a rigorous dialogue quality evaluator. Your task is to evaluate the performance of an English spoken dialogue model using a \textbf{5-point Likert scale} (1: Very Poor to 5: Excellent) and provide a brief justification for each score.

\textbf{Input Data:}
1. Dialogue Context (History); 2. User Input; 3. Model Response (Target).

\textbf{General Constraints:}
\begin{itemize}[leftmargin=*]
    \item \textbf{Format:} The response must follow: \textit{Summary/Extension + Question}. The summary must precede the question.
    \item \textbf{Style:} Natural spoken English. No written conventions (e.g., brackets for explanation, "A/B" notation).
    \item \textbf{Scoring Logic:} Start with the \textit{Base Score}, apply specific deductions/additions, and determine the final score (Min: 1, Max: 5).
\end{itemize}

\textbf{Evaluation Dimensions \& Criteria:}

\textbf{1. Topic Relevance (Base: 5, Deduction)}
\begin{itemize}[nosep]
    \item -1: Led astray by user or slight digression.
    \item -2: Does not align with topic. -3: Completely off-topic.
\end{itemize}

\textbf{2. Task Completion (Base: 3, Add/Deduct)}
\begin{itemize}[nosep]
    \item -1: No question asked, wrong length, or prompt violation.
    \item -2: Non-human symbols. -4: Chinese content.
    \item +1: Includes summary/extension. +2: Excellent extension.
\end{itemize}

\textbf{3. Information Richness (Base: 3, Addition)}
\begin{itemize}[nosep]
    \item 3: Basic question only.
    \item 4: 1--2 points of extra info. 5: $\ge$3 points of extra info.
\end{itemize}

\textbf{4. Topic Guidance (Base: 3, Addition)}
\begin{itemize}[nosep]
    \item 3: Yes/No question.
    \item 4: Fact-based question (item, color, etc.).
    \item 5: Question requiring thought/elaboration.
\end{itemize}

\textbf{Output Format:}
Output strictly in JSON format:
\texttt{\{"reason\_consistency": "...", "score\_consistency": int, ...\}}
\end{tcolorbox}
\caption{The full prompt used for LLM-based subjective evaluation. The prompt defines specific base scores and calculation rules for eight dimensions.}
\label{tab:eval_prompt}
\end{table*}
\section{Case Study: Responses from Different Models}
\label{app:case_study}

Table \ref{tab:case_study_box} illustrates a comparison of responses generated by the Base model, SFT model, and our DDPO-tuned model. The example highlights how the DDPO model adheres to vocabulary constraints (Level 3) while maintaining conversation flow, whereas baseline models introduce Out-Of-Vocabulary (OOV) terms.

\begin{table*}[h!]
\centering
\begin{tcolorbox}[
    colback=white, 
    colframe=gray!50!black, 
    title=\textbf{Case Study: Responsibility and Growth (Level 3)},
    fonttitle=\bfseries
]
\small
\textbf{Context:}
\begin{itemize}[leftmargin=4.5em, labelsep=0.5em, nosep]
    \item[\textbf{Teacher:}] Interesting observation. Do you think practice can help improve how we deal with these unexpected hurdles?
    \item[\textbf{Student:}] Yes, practice can definitely improve our ability to face and solve them.
    \item[\textbf{Teacher:}] Great point! Can you think of a time when you had to overcome a challenge that made you feel more confident and independent?
    \item[\textbf{Student:}] Yes, like when I finished a big project alone. I felt very proud and grown up.
\end{itemize}

\textbf{Model Responses:}

\vspace{0.4em}
\textbf{1. 7B Base Model} \\
\textit{``Finishing a big project alone is a huge \textbf{\textcolor{red}{accomplishment}}. How did taking on that challenge and succeeding make you feel about your own \textbf{\textcolor{red}{growth}} and \textbf{\textcolor{red}{capabilities}}?''} \\
\textbf{OOV Errors:} \textcolor{red}{\textit{accomplishment}, \textit{growth}, \textit{capabilities}} (Level $>$ L3)

\vspace{0.6em}
\textbf{2. 7B SFT Model} \\
\textit{``Proud and independent, those are wonderful feelings! How do you think this experience will influence your future \textbf{\textcolor{red}{approach}} to similar challenges?''} \\
\textbf{OOV Error:} \textcolor{red}{\textit{approach}} (Level $>$ L3)

\vspace{0.6em}
\textbf{3. 7B DDPO (Ours)} \\
\textit{``Fantastic connection! Feeling proud makes a big difference. How does feeling proud make overcoming challenges feel more successful to you?''} \\
\textbf{OOV Error:} \textcolor{green!60!black}{\textbf{None}}

\end{tcolorbox}
\caption{Comparison of model responses. The Base and SFT models fail to strictly follow the vocabulary constraints, generating OOV words (marked in red). The DDPO model successfully generates a coherent response within the L3 vocabulary limit.}
\label{tab:case_study_box}
\end{table*}

\section{Detailed Case Study: Entropy Collapse}
\label{app:entropy_collapse_case}

To provide a concrete illustration of the entropy collapse phenomenon discussed in Section 4.1, Table \ref{tab:collapse_samples} presents a comparison of multiple responses generated by the Base Model, GRPO, and DDPO given the \textbf{same dialogue context}.

We generated 8 independent samples for each model using the same temperature setting ($T=1.0$). 
\begin{itemize}
    \item \textbf{Base Model:} Shows natural diversity but lacks specific pedagogical optimization.
    \item \textbf{GRPO (Baseline):} Exhibits severe \textbf{entropy collapse}. As highlighted in \textcolor{red}{red}, the model converges to a single, repetitive pattern: it simply restates the user's input ("You drink water with your lunch") to satisfy length constraints safely, followed by a repetitive question structure. The semantic diversity is near zero.
    \item \textbf{DDPO (Ours):} Successfully maintains high diversity in phrasing (e.g., "Great", "Wonderful", "Fantastic") and sentence structure while adhering to the educational constraints.
\end{itemize}

\begin{table*}[h]
    \centering
    \small
    \renewcommand{\arraystretch}{1.3}
    \begin{tabularx}{\textwidth}{p{2cm}|X}
        \toprule
        \multicolumn{2}{c}{\textbf{Context}} \\
        \multicolumn{2}{p{0.95\textwidth}}{\textit{History:} [Tutor: "Great! What do you usually have for lunch?", Student: "Rice, vegetables, and meat.", Tutor: "Delicious! Do you drink water with your lunch?"]} \\
        \multicolumn{2}{p{0.95\textwidth}}{\textbf{User Input:} "Yes, I drink water with lunch."} \\
        \midrule
        \midrule
        
        \textbf{Model} & \textbf{Sampled Responses (8 independent generations)} \\
        \midrule
        
        \textbf{Base Model} & 
        1. Good habit! Do you like eating fruits after lunch? \newline
        2. Good habit! Do you eat rice for breakfast too? \newline
        3. Good for you! Do you eat rice, vegetables, and meat for dinner too? \newline
        4. Good habit! What's your favorite meat to eat with rice and vegetables? \newline
        5. Good habit! Do you like eating fruit after lunch? \newline
        6. Good habit! Do you like eating fruits after lunch? \newline
        7. Good habit! Do you like eating fruits with your lunch too? \newline
        8. Good for you! Do you like eating fruits with your meals?
        \\ \hline
        
        \textbf{GRPO} & 
        \textcolor{red}{1. You drink water with your lunch.} Do you eat rice with your lunch? \newline
        \textcolor{red}{2. You drink water with your lunch.} Do you eat rice with your lunch? \newline
        \textcolor{red}{3. You drink water with your lunch!} Do you eat meat for lunch? \newline
        \textcolor{red}{4. You drink water with your lunch!} Do you eat rice with your lunch? \newline
        \textcolor{red}{5. You drink water with your lunch!} Do you eat rice with your lunch? \newline
        \textcolor{red}{6. You drink water with your lunch!} Do you eat meat with your lunch? \newline
        \textcolor{red}{7. You drink water with your lunch.} Do you eat rice for lunch? \newline
        \textcolor{red}{8. You drink water with your lunch!} Do you eat rice with your lunch?
        \\ \hline
        
        \textbf{DDPO (Ours)} & 
        1. \textbf{Great!} Eating vegetables makes food taste nice! What kind of food do you like? \newline
        2. \textbf{Wonderful!} Water makes eating feel good! What kind of drinks do you have? \newline
        3. \textbf{Wonderful!} Water helps make food taste good! What makes food so nice? \newline
        4. \textbf{Great!} Water makes eating feel good! What kind of drinks do you have? \newline
        5. \textbf{Wonderful!} Water makes eating very nice! What kinds of drinks do you have? \newline
        6. \textbf{Great!} Water helps make food taste better! What kind of food do you like? \newline
        7. \textbf{Fantastic!} Water makes eating feel good! What kind of food do you like? \newline
        8. \textbf{Great!} Water makes eating very nice! What kind of drinks do you have?
        \\ 
        \bottomrule
    \end{tabularx}
    \caption{Comparison of generation diversity. The GRPO baseline collapses into a repetitive template (repeating user input + simple question), whereas DDPO generates diverse feedback and follow-up questions.}
    \label{tab:collapse_samples}
\end{table*}

\section{Use of AI Tools}
We utilized AI-based tools (e.g., ChatGPT, Gemini) to assist in refining the clarity, grammar, and readability of the manuscript. The conceptualization, experimental design, data analysis, and scientific conclusions remain the sole work and responsibility of the authors. All AI-suggested modifications were reviewed and verified by the authors.

\end{document}